\pdfoutput=1

\documentclass[11pt]{article}
\usepackage[hyperref]{acl2021}
\usepackage[T1]{fontenc}
\usepackage[utf8]{inputenc}
\usepackage{microtype}
\usepackage{times}
\usepackage{enumitem}
\usepackage{latexsym}
\usepackage{graphicx}
\usepackage{tabularx}
\usepackage{threeparttable}  
\usepackage{booktabs} 
\usepackage{amsmath}
\usepackage{url}
\usepackage{amsfonts,amssymb}
\usepackage{multirow}
\usepackage{multicol}

\usepackage{microtype}

\usepackage{adjustbox}

\def\eg{\emph{e.g.}}

\newcommand{\banking}{\text{\small BANKING}77 }
\newcommand{\hwu}{\text{\small HWU}64 }
\newcommand{\clinic}{\text{\small CLINIC}150 }
\newcommand{\restaurant}{\text{\small REST}8\text{\small K} }
\newcommand{\multiWOZ}{M\text{\small ULTI}WOZ }

\title{SPACE-2: Tree-Structured Semi-Supervised Contrastive Pre-training \\ for Task-Oriented Dialog Understanding}

\author{
Wanwei He\textsuperscript{1,2, \includegraphics[scale=0.02]{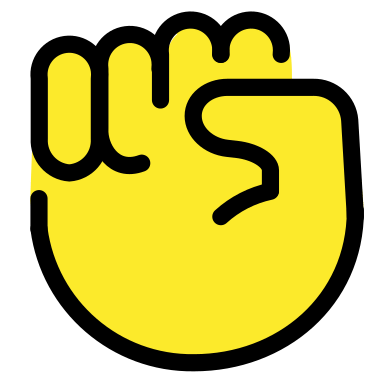}}, 
Yinpei Dai\textsuperscript{3,\includegraphics[scale=0.02]{first.png}}, 
Binyuan Hui\textsuperscript{3,\includegraphics[scale=0.02]{first.png}}, 
Min Yang\textsuperscript{1}\thanks{\quad Corresponding authors.}, 
Zheng Cao\textsuperscript{3},
\\
\bf{
Jianbo Dong\textsuperscript{3}, 
Fei Huang\textsuperscript{3}, 
Luo Si\textsuperscript{3}, 
Yongbin Li\textsuperscript{3}\footnotemark[1]
}
\\
\textsuperscript{1}Shenzhen Institute of Advanced Technology, Chinese Academy of Sciences, China\\
\textsuperscript{2}University of Chinese Academy of Sciences, China\\
\textsuperscript{3}DAMO Academy, Alibaba Group\\
\texttt{\{ww.he, min.yang\}@siat.ac.cn} \\
\texttt{\{yinpei.dyp, f.huang, luo.si, shuide.lyb\}@alibaba-inc.com}
}

\date{}

\begin{document}
\maketitle
\renewcommand{\thefootnote}{\fnsymbol{footnote}}
\footnotetext{\includegraphics[scale=0.02]{first.png} Equal contribution.}
\begin{abstract}
Pre-training methods with contrastive learning objectives have shown remarkable success in dialog understanding tasks. However, current contrastive learning solely considers the self-augmented dialog samples as positive samples and treats all other dialog samples as negative ones, which enforces dissimilar representations even for dialogs that are semantically related. In this paper, we propose \textbf{SPACE-2}, a tree-structured pre-trained conversation model, which learns dialog representations from limited labeled dialogs and large-scale unlabeled dialog corpora via semi-supervised contrastive pre-training. Concretely, we first define a general semantic tree structure (STS) to unify the inconsistent annotation schema across different dialog datasets, so that the rich structural information stored in all labeled data can be exploited. Then we propose a novel multi-view score function to increase the relevance of all possible dialogs that share similar STSs and only push away other completely different dialogs during supervised contrastive pre-training. To fully exploit unlabeled dialogs, a basic self-supervised contrastive loss is also added to refine the learned representations. Experiments show that our method can achieve new state-of-the-art results on the DialoGLUE benchmark consisting of seven datasets and four popular dialog understanding tasks. For reproducibility, we release the code and data at \url{https://github.com/AlibabaResearch/DAMO-ConvAI/tree/main/space-2}.
\end{abstract}

\section{Introduction}
\label{sec:intro}
Task-oriented dialog (TOD) systems aim to help users to accomplish specific tasks through natural language interactions, \eg, restaurant booking, hotel reserving, and movie searching \cite{young2013pomdp}. 
In order to fulfill a user goal, the dialog systems must be capable of extracting structured semantics from the dialog utterances, which is often referred to as \textit{dialog understanding}. 
Common tasks of dialog understanding include: 
1) intent prediction, where the system classifies the user intention given the current utterance,
2) slot filling, where the system extracts the value spans for pre-defined slots from user utterances,
and 3) dialog state tracking, where slot-value pairs are predicted and updated according to multi-turn dialog history.

Recent methods address the dialog understanding problems mainly by task-adaptive training with pre-trained language models (PLMs) \cite{henderson-etal-2020-convert, mehri2020dialoglue, zhang-etal-2022-slot}.
By fine-tuning the entire parameters of PLMs on dialog corpora, the versatile knowledge acquired from large-scale corpora can be adapted to handle various dialog understanding tasks.
For example, TOD-BERT \cite{wu-etal-2020-tod} continually pre-trained a BERT on large-scale task-oriented dialog corpora with similar objectives to grasp more knowledge in conversations and achieved good results on a wide range of tasks.
To further enhance the representations of dialogs, contrastive learning is widely used for dialog pre-training via pulling semantically similar sentences together and pushing apart dis-similar ones, which has shown strong ability in few-shot dialog understanding \cite{zhang-etal-2020-discriminative, zhang2021few}.

\begin{figure}[t]
    \centering
    \includegraphics[width=0.45\textwidth]{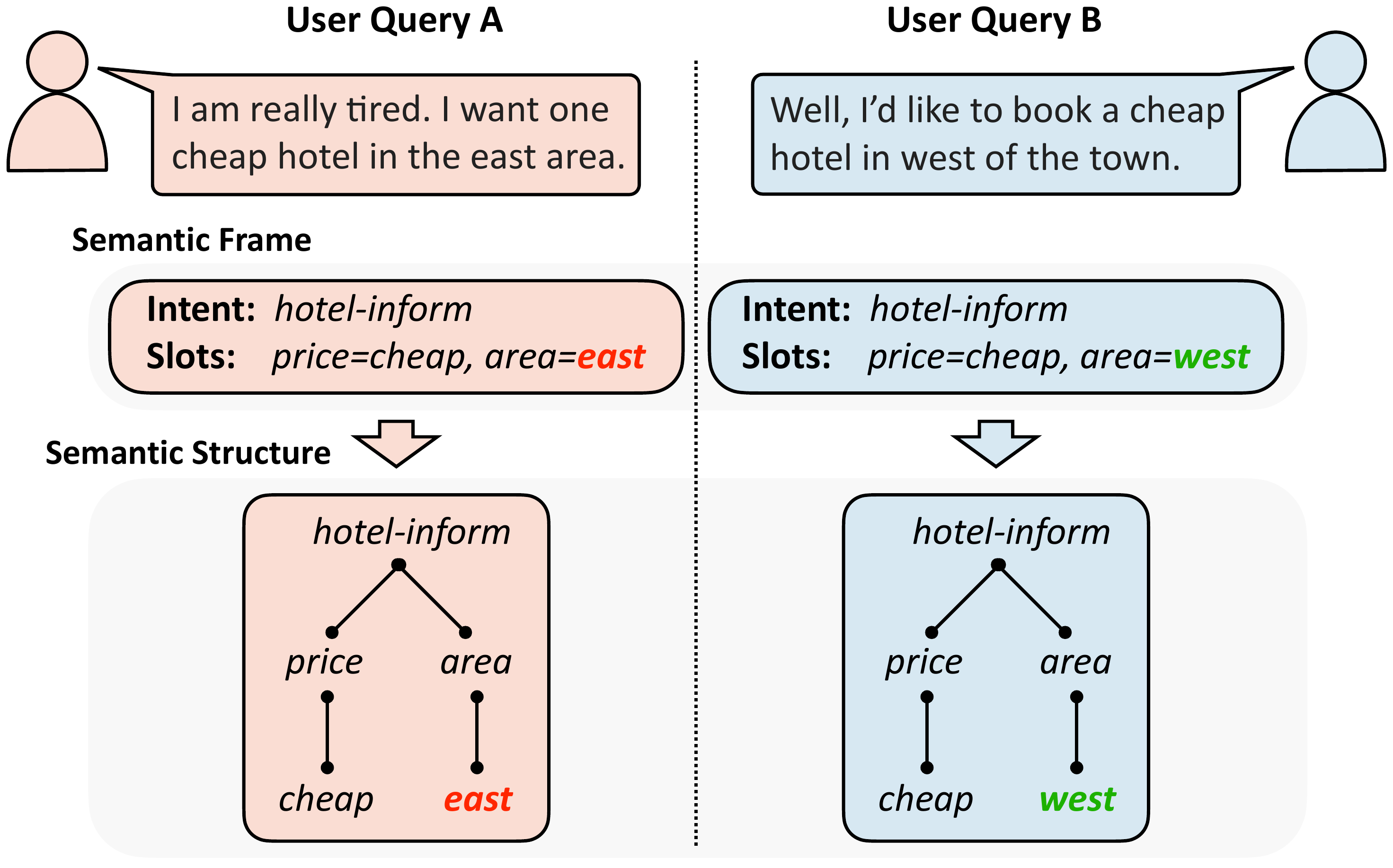}
    \caption{Two dialog examples that have similar semantic contents from the \multiWOZ dataset. The difference is highlighted in red and green colors.}
    \label{fig:example}
\end{figure}

Although contrastive learning (CL) is effective in pre-training dialog models, it still suffers from several problems. 
First, the current practice of CL often neglects the semantic structures of dialog examples and only utilizes shallow  self-supervised information.
Figure \ref{fig:example} illustrates the semantic frames of two dialog examples, which indicate the hierarchical  structures with semantic meanings in terms of intentions and slots. The semantic structure is often summarized as the turn-level structural annotation and is prominent in multi-domain task-oriented dialogs.
Exploiting such structure during task-adaptive pre-training has been proved to be effective in dialog understanding tasks \cite{yu2020score}. However, 
In conventional CL, only the examples with the exact same annotations are deemed as positive samples, while all other examples are considered as negative ones \cite{gao-etal-2021-simcse}, ignoring the fact that different examples may share similar semantics to some extent as shown in Figure \ref{fig:example}.
Second, in previous dialog pre-training methods, no annotations have been exploited to learn better pre-trained representations. But in computer vision, there have already been  many works demonstrating that labeled data can accelerate the pre-training procedure as well as improve the model performance \cite{assran2020supervision, khosla2020supervised, dai2021unimoco}. 
Therefore, we argue that it is crucial to combine both the labeled and unlabeled dialog data to learn more powerful pre-trained dialog models.

Built on these motivations, we propose a novel tree-structured pre-trained conversation model (SPACE-2), which fully exploits the structural information in labeled data to improve the dialog pre-training on large-scale unlabeled corpora via semi-supervised contrastive learning.
To be specific, we first define a general semantic tree structure (STS), which is suitable for TOD and compatible with the discrepancy of annotation schema in varied dialog datasets. 
Then we propose a novel multi-view scoring function to measure the similarity among different labeled dialogs. Given two dialogs with STSs, the scoring function is able to calculate their relations hierarchically and aggregate the output multiple scores from either global or local aspects.
In this way, all labeled data are viewed as positive samples with soft scores instead of hard scores (0 or 1) in traditional CL. Thus, more subtle semantic structures in training samples can be taken into account,
and dialog representations can be learned in a semi-supervised manner.

Our contributions are summarized as follows:
\begin{itemize}
    \item We explore tree-structured semi-supervised contrastive pre-training for task-oriented dialog understanding.
    To the best of our knowledge, this is the first study to inject structural information and exploit labeled semantic frames in a pre-trained conversation model. 
    \item We propose a multi-view score function for similarity measurement among different labeled dialogs, which could calculate relations hierarchically and aggregate multiple judgements from either global or local aspects.
    \item We conduct extensive experiments on the DialoGLUE benchmark with seven different datasets. Empirical results show that our SPACE-2 consistently performs better than strong competitors on four dialog understanding tasks (i.e., intent prediction, slot filling, semantic parsing, and dialog state tracking).  
\end{itemize}









%

\section{Related Work}
\label{sec:relate}


\subsection{Pre-trained Models for Dialog Understanding}
Recent advances in pre-trained language models (PLMs) have spurred success in natural language understanding in task-oriented dialog (TOD) systems. Since general-purpose PLMs such as BERT \cite{devlin-etal-2019-bert} are not able to capture dialog-oriented knowledge in TOD, such as explicit user intent, many studies proposed to perform pre-training on dialog corpora. For example, ConvRT \cite{henderson-etal-2020-convert} explored the pre-trained language transformers for TOD via learning a dialog response selection loss. 
ConvBERT \cite{mehri2020dialoglue}
fine-tuned BERT on a
large open-domain dialog corpus using a masked language modeling objective so that more semantically meaningful representations of dialogs can be learned.
TOD-BERT \cite{wu-etal-2020-tod}  incorporated role tokens in the input and employed a contrastive objective for response selection. 
\citet{zhang2021dialoguebert} proposed to use masked utterance modeling and response contrastive loss to produce  DialogueBERT and showed good performance on intent, emotion, and entity recognition.
Apart from training with common objectives, some work \cite{yu2020score, he2020amalgamating, hui2021dynamic, dai-etal-2021-preview} also proposed to learn relations between schema and keywords of utterances during pre-training, which benefits in certain downstream tasks that require parsing semantics. The work of \cite{sun2019ernie, liu2020k} tried to integrate structural information of knowledge graphs to enhance language representations. 
Unlike these methods, we are the first to inject the semantic tree structure into pre-training models for dialog understanding. 

\begin{table}[t]
\centering
\scalebox{0.8}{
\begin{tabular}{cccc}
\toprule
Statistics & \textit{AnPreDial} & \textit{UnPreDial}  \\
\midrule
\# Datasets & 32 & 21  \\
\# Dialogs & 459,465 & 3,217,058 \\
\# Turns & 3,366,479 & 19,578,028 \\
Avg. tokens per turn & 13.9 & 14.5  \\
Avg. tokens per dialog & 101.8 & 88.2  \\
Total unique tokens & 46.8M & 283.7M \\
\bottomrule
\end{tabular}}
\caption{Statistics of our pre-trained dialog dataset.}
\label{tab:dataset}
\end{table}

\subsection{Contrastive Learning}
Contrastive learning (CL) is one of the pre-training methods that leverage large-scale unlabeled data to learn meaningful sentence representations.
The key idea of CL is to narrow the distance between two semantically similar sentence representations, meanwhile,  push away the representations of dissimilar sentences \cite{gao-etal-2021-simcse, wu2021esimcse, yan2021consert}. There are also several studies focused on applying CL to dialog understanding tasks. For example, \citet{zhang2021few} used CL as self-supervised pre-training and adapted the pre-trained models on few-shot intent prediction. The studies \cite{mehri-eric-2021-example, vulic2021convfit} utilized supervised CL at the fine-tuning stages for intent prediction, where the samples from the same class are all regarded as positives.  \citet{he2020contrastive} proposed to use  adversarial samples as difficult negative instances in CL for zero-shot cross-domain slot-filling.  
Different from previous work, we use CL in both pre-training and fine-tuning stages. In addition, we train dialog models with  both labeled and unlabeled data in a semi-supervised manner. 

\subsection{Semi-Supervised Contrastive Learning}
Semi-supervised contrastive learning has proved to be an effective pre-training method in many research field, such as image recognition \cite{yuan2021activematch, park2021opencos, li2021comatch, kim2021selfmatch}, image segmentation \cite{alonso2021semi, zhou2021c3} and speech recognition \cite{xiao2021contrastive, inoue2020semi}. It can combine both labeled and unlabeled data together to train effective neural models for downstream tasks. 
\citet{he2022galaxy} proposed to use consistency regularization loss to learn dialog policy from labeled and unlabeled dialog corpora via a semi-supervised manner.
However, leveraging such a semi-supervised contrastive learning paradigm in task-oriented dialog understanding remains unexplored. In this paper, we explore semantic structures in TOD, and propose a new contrastive learning approach that calculates semantic tree-structure similarity among all possible labeled dialog data. 

\section{Pre-training Data}
In this section, we first describe the dialog datasets \cite{he2022unified} used for pre-training, including a labeled dialog dataset (\textit{AnPreDial}) and a large-scale unlabeled dialog corpus
(\textit{UnPreDial}).
Then, we elaborate on the semantic tree structure (STS) for the task-oriented dialog data.

\subsection{Dataset Description}
To provide sufficient high-quality dialog corpora to train our pre-trained conversation model, we use a labeled dialog dataset called \textit{AnPreDial}, which contains 32 existing labeled TOD datasets, ranging from single-turn question answering to multi-turn dialogs. We also use a large-scale unlabeled dialog corpus called \textit{UnPreDial} with 21 dialog datasets, ranging from open-domain online forums to document-grounded dialogs. The statistics of our pre-trained dialog corpora are shown in Table \ref{tab:dataset} and more details can be found in Appendix \ref{sec: appen}.

\subsection{Semantic Tree Structure}
Task-oriented dialogs often contain several user goals, and the system needs to help fulfill these goals through interacting with users. To this end, the user utterances and system responses are usually composed of rich semantics such as intents and slots. In different datasets, the annotation taxonomy used to describe the semantic scopes varies. For example, in some intent recognition datasets such as \banking \cite{casanueva-etal-2020-efficient} and \text{\small CLINIC}150  \cite{larson-etal-2019-evaluation}, they only possess a pre-defined intent set and formulate the task as a classification problem. In other datasets like Taskmaster \cite{byrne2019taskmaster}, only the slot labels are provided; for instance, given a user utterance \textit{``I need a table for tonight at 7 pm for 8 people"}, the label is \textit{``time.reservation=7 pm, num.guests=8"} without any intent information.
Therefore, in order to utilize all possible data in \textit{AnPreDial} to pre-train our model, we adopt a unified schema called semantic tree structure (STS) that is suitable for every TOD sample.

\begin{figure}[t]
    \centering
    \includegraphics[width=0.45\textwidth]{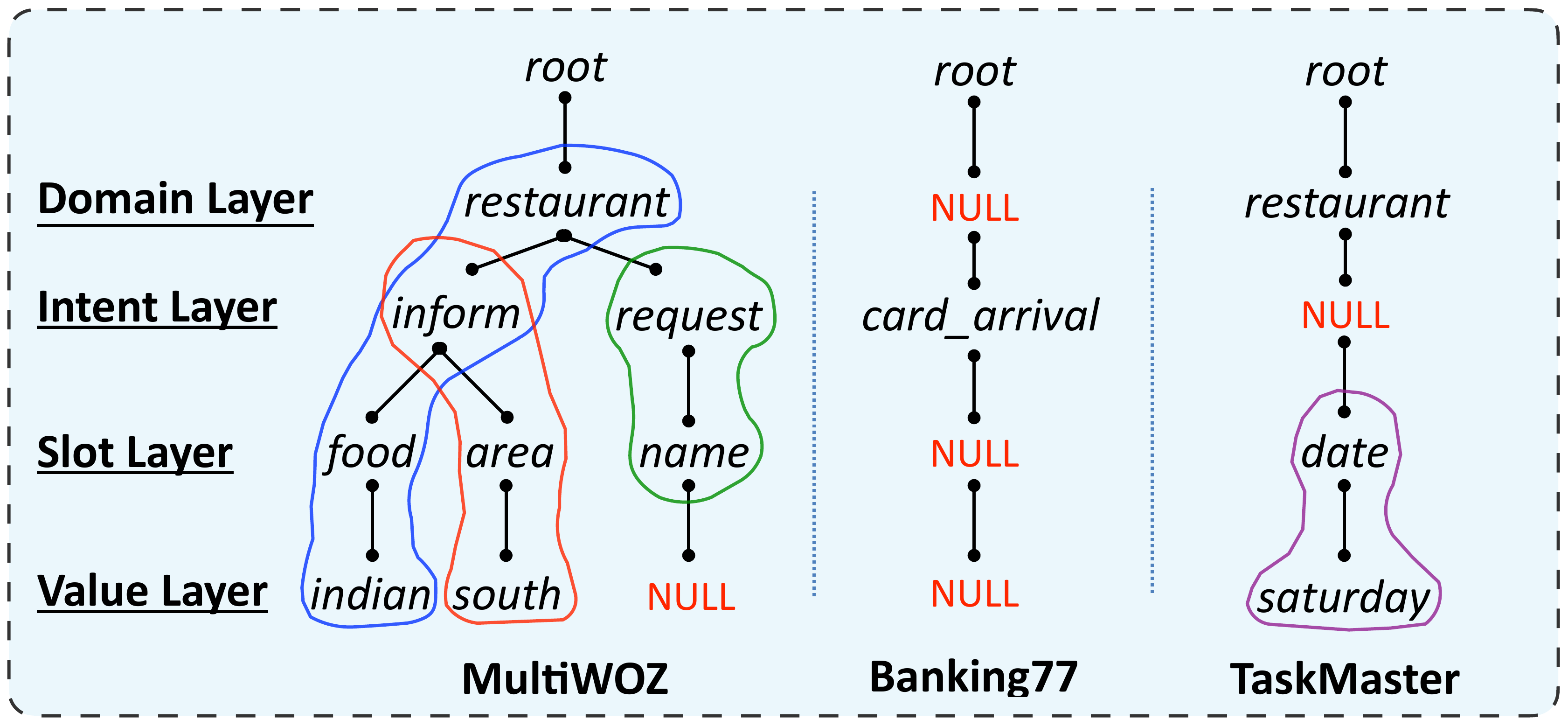}
    \caption{Different semantic tree structures in \textit{AnPreDial}, where the blue, red, green and purple circles indicate \texttt{DISV}, \texttt{ISV}, \texttt{IS}, \texttt{SV} sub-structures.}
    \label{fig:tree_type}
\end{figure}

Typically, the STS contains four layers: \textit{domain} layer, \textit{intent} layer, \textit{slot} layer, and \textit{value} layer.
Every layer is composed of nodes, which are referred to as respective elements of labeled semantic frames.
If no matched annotations are provided, the nodes of the corresponding layer are set as empty (denoted as \texttt{NULL}). 
The first layer consists of domain nodes as successors of the root node, preceded by the intent nodes.
As children of intent nodes, slot nodes occupy the third layer of the semantic tree.
As leaf nodes, the value nodes take the slot nodes as parents.
Figure \ref{fig:tree_type} shows three tree structures of different dialog samples from \multiWOZ \cite{eric-etal-2020-multiwoz}, \banking and Taskmaster.
In \multiWOZ, the labels follow the  dialog act schema in \citet{young2007cued}, and all layers contain certain semantic elements. But the other two datasets do not follow the schema and lack respective elements in some layers.

\section{Methods}
\label{sec:method}


In this section, we first introduce the model architecture.
Then we expound on the pre-training procedure with the proposed tree-structured semi-supervised contrastive learning (CL) paradigm.
\begin{figure}[t]
    \centering
    \includegraphics[width=0.4\textwidth]{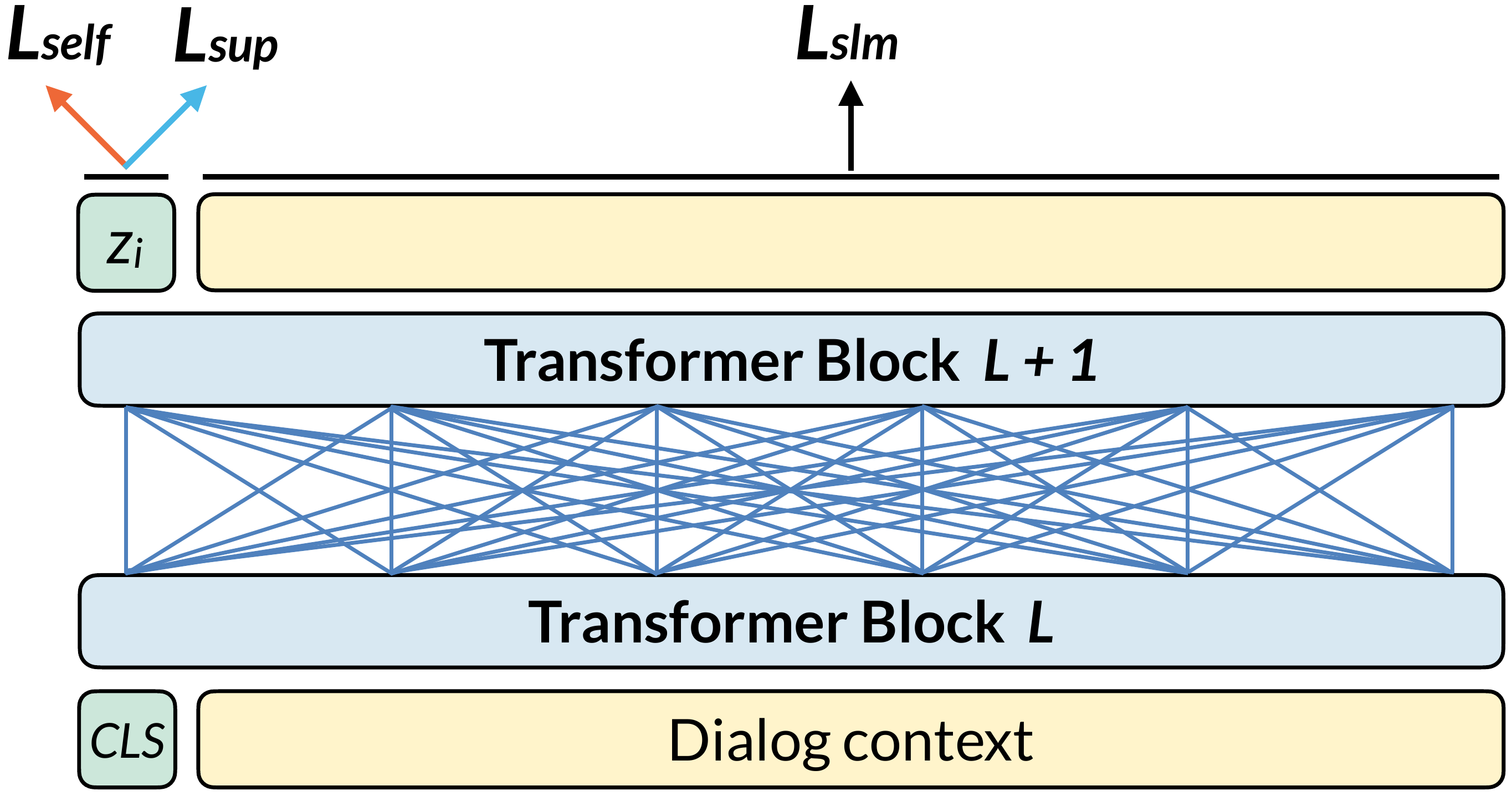}
    \caption{Model Framework. $L_{self}$ is the tree-structured self-supervised contrastive loss for unlabeled data, and $L_{sup}$ is the tree-structured supervised contrastive loss for labeled data. $L_{slm}$ is the span MLM loss for all data.}
    \label{fig:model}
\end{figure}

\begin{figure}[t]
    \centering
    \includegraphics[width=0.48\textwidth]{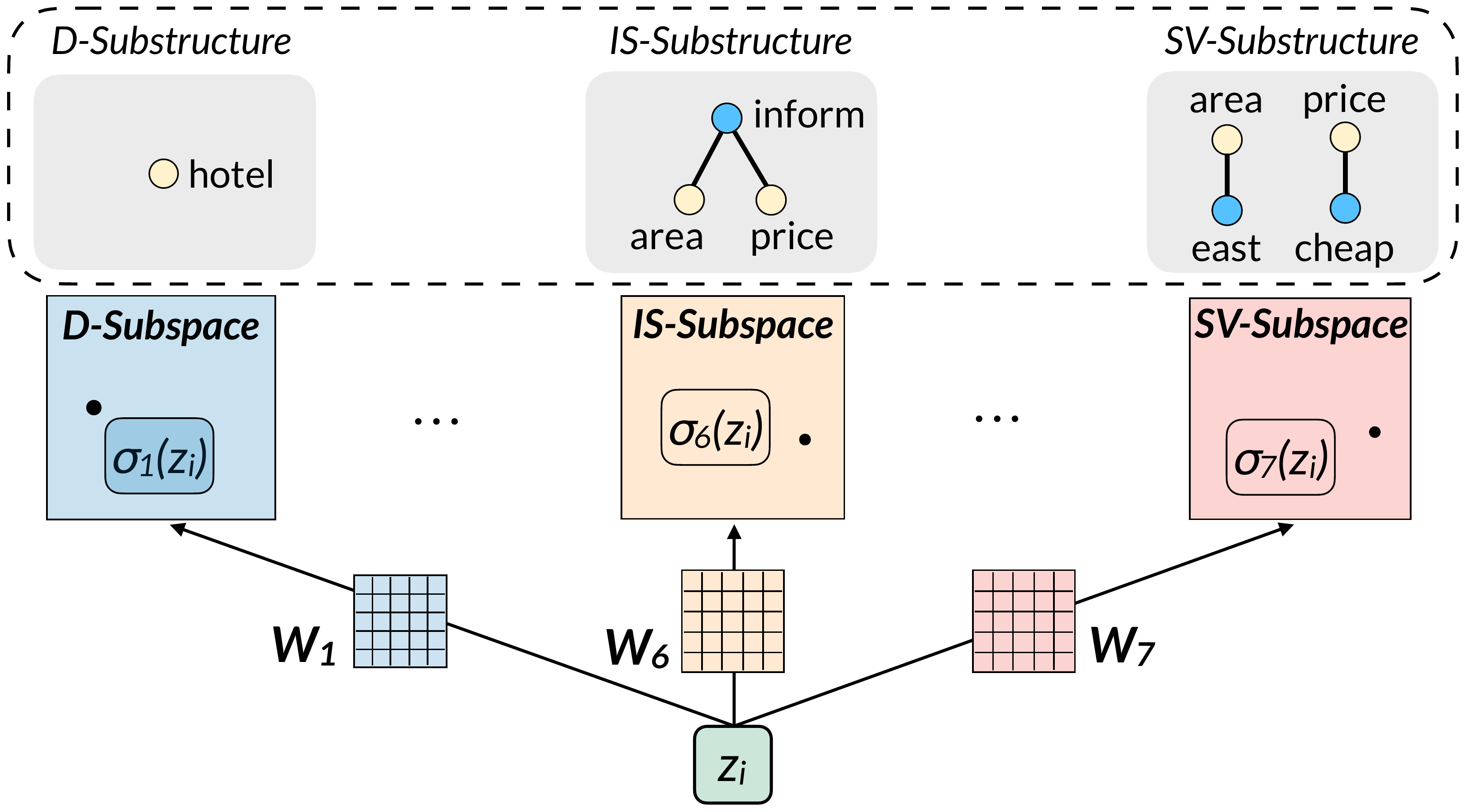}
    \caption{Sub-spaces for the multi-view scoring objective. The dashed black line denotes the annotations of different sub-structures, only used in our supervised CL.}
    \label{fig:multiview}
\end{figure}

\subsection{Model Architecture}
As illustrated in Figure \ref{fig:model}, we build our SPACE-2 model based on the bidirectional Transformer architecture \cite{vaswani2017attention}. 
Different from the vanilla input representations as in BERT \cite{devlin-etal-2019-bert}, we set our input embeddings consisting of four elements: tokens, roles, turns, and positions. Role embeddings are used to segment which role the current token belongs to either user or system. Turn embeddings are assigned to each token according to its turn number in the dialog. Position embeddings are assigned to each token according to its relative position within its belonging sentence. Appendix \ref{sec: input_representations} shows more details. 


\subsection{Tree-Structured Semi-Supervised Contrastive Learning Paradigm}
\label{sec:scl}

We aim to leverage semi-supervised pre-training to learn better pre-trained representations from both the labeled and unlabeled data. 
Concretely, we adopt a tree-structured supervised contrastive objective on the labeled dataset \textit{AnPreDial}, while a tree-structured self-supervised contrastive objective on the unlabeled dataset \textit{UnPreDial}. 

\subsubsection{Tree-Structured Supervised CL}
For labeled data, we treat every dialog sample as a positive sample assigned with similarity scores to each other. Between two semantic tree structures (STSs), we define a score function to calculate similarity scores from different views. Let \texttt{D}, \texttt{I}, \texttt{S}, \texttt{V} denote the sets of node in domain, intent, slot and value layer.
For the sample from \multiWOZ dataset in Figure \ref{fig:tree_type}, its \texttt{D}=\{\textit{restaurant}\}, \texttt{I}=\{\textit{inform}, \textit{request}\}, \texttt{S}=\{\textit{food, area, name}\} and \texttt{V}=\{\textit{indian, south}\}.
For the sample from \banking dataset, its \texttt{D}=\{\}, \texttt{I}=\{\textit{card\_arrival}\}, \texttt{S}=\{\} and \texttt{V}=\{\}. 
We also build the sets of each possible path between two nodes on the STS, which we denote as \texttt{DI}, \texttt{IS}, \texttt{SV}, \texttt{DIS}, \texttt{ISV}, \texttt{DISV}. 
Thus for the \multiWOZ sample in Figure \ref{fig:tree_type}, \texttt{ISV}=\{\textit{inform$\rightarrow$food$\rightarrow$indian, inform$\rightarrow$area$\rightarrow$south}\}, and for \banking,  \texttt{ISV}=\{\}. 
The total super set of all above node-sets and path-sets is denoted as $\mathcal{S}_{all}=\{\texttt{D}, \texttt{I}, \texttt{S}, \texttt{V}, \texttt{DI}, \texttt{IS}, \texttt{SV}, \texttt{DIS}, \texttt{ISV}, \texttt{DISV}\}$, where the size $K=10$.
Given a pair of STSs (denoted as $T_i,T_j$), then we calculate the Jaccard similarity coefficient for each set $s_k$ in $\mathcal{S}_{all}$ to get the score $f_k$ respectively as follows: 
\begin{gather}
    f_{k}(T_i, T_j) = \mathrm{J}(s_k^{i}, s_k^{j})\\
    \mathrm{J}(A, B)= \frac{|A\cap B |}{|A\cup B|}
\end{gather}
where we use $i, j$ to discriminate $s_k$ and $\mathcal{S}_{all}$ for two different STSs: $s_k^{i} \in \mathcal{S}_{all}^{i}, s_k^{j} \in \mathcal{S}_{all}^{j}$. $\mathrm{J}$ is the function for the Jaccard similarity coefficient, which takes the ratio of Intersection over Union. 

Given the dialog context, our model output a pooled representation at the \texttt{[CLS]} position as the sentence embedding $z$ of the whole context.
Thus for any sample pair $i, j$, the output sentence embeddings are denoted as $z_i, z_j$. We use $f_{i,j}^k$ to denote the $k$-th similarity score $f_k(T_i, T_j)$ for simplicity. Suppose the size of the current batch is $N$, we follow the common practice in \citet{gao-etal-2021-simcse} to duplicate the data via dropout-based data augmentation. Then we acquire a $2N$-length new batch. Let $I=\{1\dots2N\}$ be the index set of the new batch, and  $C(i) \equiv I / \{i\}$. Since we compute multiple scores $f_{i,j}^k$ for each STS pair, different types of scoring functions can be used to construct the CL loss. In this paper, We propose both single-view and multi-view scoring functions to build the tree-structured supervised CL objectives as follows and find that the multi-view method performs better.

\noindent\textbf{Single-View Scoring Objective.} Similar to the common practice in the current CL, we simply average the $K$ scores into a single value to weigh the semantic similarity among samples:
\begin{align}
L_{sup}^{single}& = - \sum_{i \in I} \sum_{j \in C(i)} \left(\frac{1}{K}\sum_{k=1}^{K} f^k_{i, j}\right) \times \\ \nonumber
& \log \frac{\exp \left(\sigma({z}_{i})\cdot \sigma({z}_{j}) / \tau\right)}{\sum_{l \in C(i)} \exp \left(\sigma({z}_{i}) \cdot \sigma({z}_{l})/ \tau\right)}
\end{align}
\begin{equation}
\sigma(z) = \mathrm{Norm}(W z + b)
\end{equation}
where $\sigma(\cdot)$ is a normalized linear mapping to transform $z$ to a unit vector. $W$ is a learnable square weight matrix and $b$ is a bias vector.  $\mathrm{Norm(\cdot)}$ is the normalization operation and $\tau\in[0,1]$ is a temperature hyper-parameter.

\begin{figure*}[htp]
    \centering
    \includegraphics[width=1.0\textwidth]{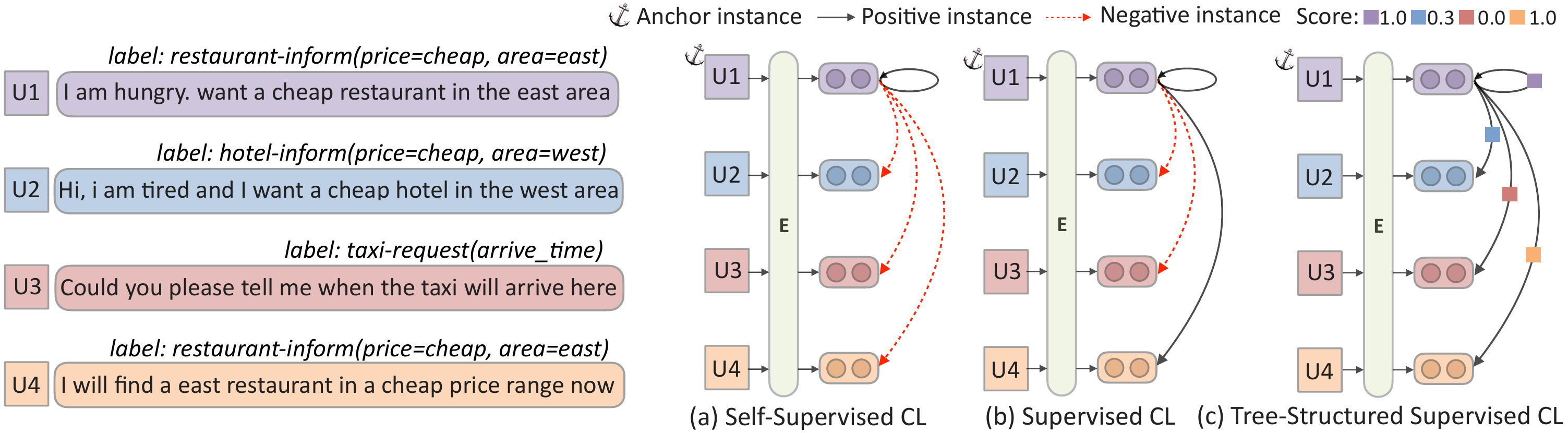}
    \caption{(a) Self-supervised CL only predicts augmented itself from in-batch negatives, with different dropout masks applied. (b) Supervised CL considers samples of the exact same label as positives.  (c) Tree-structured supervised CL considers all in-batch samples as positives with soft scores. Only scores of \texttt{ISV} set are depicted here.}
    \label{fig:case}
\end{figure*}

\noindent\textbf{Multi-View Scoring Objective.} Figure \ref{fig:multiview} demonstrates our multi-view scoring objective. Instead of summing up as one score, we can separately utilize multiple scores to better exploit sub-structures from different semantic spaces \cite{zhu2019multi} as the following loss:
\begin{align}
L_{sup}^{multi}& = - \sum_{i \in I} \sum_{j \in C(i)} \sum_{k=1}^{K} \left( \frac{f^k_{i, j}}{\sum_{m \in C(i)} f^k_{i, m}} \times \right.\\ \nonumber
& \left.\log \frac{\exp \left(\sigma_k({z}_{i})\cdot \sigma_k({z}_{j}) / \tau\right)}{\sum_{l \in C(i)} \exp \left(\sigma_k({z}_{i}) \cdot \sigma_k({z}_{l})/ \tau\right)}\right)
\end{align}
\begin{equation}
\sigma_k(z) = \mathrm{Norm}(W_k z + b_k)
\end{equation}
where $\sigma_k(\cdot)$ denotes linear mappings with different parameters.
As illustrated in Figure \ref{fig:case}, our methods are quite distinct from vanilla supervised CL \cite{vulic2021convfit}, where only augmented data or data from the same label are considered as positive samples ($f^k_{i,j} = 1$), and all other data are considered as negative samples ($f^k_{i,j} = 0$). 
Therefore, our method can be viewed as a generalized case where all labeled data are regarded as positive samples assigned with soft scores.

\subsubsection{Tree-Structured Semi-Supervised CL}
For unlabeled data, since there are no available labels, we adopt a tree-structured self-supervised contrastive objective in a similar way in \citet{gao-etal-2021-simcse}, where only the augmented data by dropout is deemed as a positive sample.
In particular, under the multi-view setting, the objective $L^{multi}_{self}$ still shares the same $K$ sub-spaces as $L^{multi}_{sup}$ as follows:
\begin{align}
L_{self}^{single} &= - \sum_{i \in I} \sum_{j \in C(i)} \textbf{1}(j=i^+) \times \\ \nonumber
&  \log \frac{
\exp \left(\sigma({z}_{i})\cdot \sigma({z}_{j}) / \tau\right)}{\sum_{l \in C(i)} \exp \left(\sigma({z}_{i})\cdot \sigma({z}_{l}) / \tau\right)}
\end{align}
\begin{align}
L_{self}^{multi} &= - \sum_{i \in I} \sum_{j \in C(i)} \sum_{k=1}^{K} \textbf{1}(j=i^+) \times \\ \nonumber
&  \log \frac{
\exp \left(\sigma^k({z}_{i})\cdot \sigma^k({z}_{j}) / \tau\right)}{\sum_{l \in C(i)} \exp \left(\sigma^k({z}_{i})\cdot \sigma^k({z}_{l}) / \tau\right)}
\end{align}
where $\textbf{1}(j=i^+)$ means it only takes 1 when the data sample $j$ is the augmented data $i^+$ of data sample $i$, otherwise 0. 


In order to maintain the language modeling ability and extract the slot values better, we also apply the spanMLM loss $L_{slm}$ \cite{joshi-etal-2020-spanbert} to the output representations on top of the dialog context. 
Therefore, our final loss is computed as:
\begin{equation}
    L_{total}^{*} = L_{sup}^{*} + L_{self}^{*} + L_{slm}
\end{equation}
where $^{*}$ can be either \textit{single} or \textit{multi}.




\section{Experimental Setup}
\label{sec:exp}

\subsection{Task Formulation}
DialoGLUE benchmark consists of seven different datasets, including four tasks as follows:

\noindent\textbf{Intent Prediction.} This is a classification task for models to predict which intent class the user conveys in the utterance.
There are three datasets accordingly: 1) \banking  \cite{casanueva-etal-2020-efficient}, a single turn dataset in the banking domain with 77 intent labels and 13k utterances; 2) \clinic  \cite{larson-etal-2019-evaluation}, a multi-domain dataset that contains 23.7k utterances, 150 intents, 10 domains; and 3) \hwu \cite{liu2021benchmarking} includes 25.7k utterances for 64 intents spanning 21 domains. To conduct the downstream task, a linear layer neural network is added to the pooled output $z$ to classify. The evaluation metric is accuracy.


\noindent\textbf{Slot Filling.} This is a sequential labeling task in which a model tries to predict the IOB tags upon the sequence of dialog context. There are two datasets: 1) \restaurant  \cite{coope-etal-2020-span}, which contains 8.2k utterances and 5 slots (date, time, people, first name, last name) from a restaurant domain; 2) a transformed DSTC8  \cite{rastogi2020towards} dataset comprising of 5.5k utterances and slot annotations in 4 domains (buses, events, homes, rental cars). Both datasets are single-turn formed and every token in the utterance is predicted as either being the beginning of a slot value (B-), inside a slot value (I-) or not belonging to a slot value (O). A linear layer neural network is added to the representation of the final output layer to predict correct tags (e.g., “B-time”, “I-people”). The evaluation metric is the macro-averaged F1 score.

\noindent\textbf{Semantic Parsing.} The original TOP \cite{gupta-etal-2018-semantic-parsing} dataset targeting at hierarchical parsing is transformed into a joint task of intent prediction and slot-filling, which contains 44k utterances. The metric is  the exact-match as in  \citet{mehri2020dialoglue}.

\noindent\textbf{Dialog State Tracking.} This task is multi-turn slot-filling, where the model interprets the user meanings during the course of a dialog to maintain a dialog state in terms of slot-value pairs. \multiWOZ2.1 \cite{eric-etal-2020-multiwoz} is the chosen dataset version with 10k dialogs. Following \citet{mehri2020dialoglue}, we report joint goal accuracy and use TripPy \cite{heck-etal-2020-trippy} as the downstream model.


\begin{table*}[t]
\centering 
\resizebox{1.0\textwidth}{!}{
\begin{tabular}{lcccccccc}
\toprule
\multicolumn{9}{c}{Full data training} \\ \hline
Model & \textit{average} & \banking & \hwu & \clinic &   \restaurant   & DSTC8  & TOP & \multiWOZ \\ \hline
BERT$^\dagger$ & 86.08 & 93.02 & 89.87 & 95.93 & 95.53 & 90.05 & 81.90 & 56.30 \\
ConvBERT$^\dagger$ & 86.01 &92.95& 90.43& 97.07 &95.90& 87.58 &82.13 &56.00 \\
ConvBERT+MLM$^\dagger$ & 86.89 & 93.44 & 92.38 & 97.11 & 95.44 & 91.20 & 82.08 & 56.56 \\
ConvBERT-DG$^\dagger$ & 82.90 & 93.21& 91.64& 96.96& 93.44& 74.54 &72.22 &58.57 \\
ConvBERT-DG+MLM$^\dagger$ & 85.34 & 92.99 & 91.82 & 97.11 & 94.34 & 86.49 & 76.36 & 58.29 \\
TOD-BERT$_{adapt}$   & 86.95 & 93.61 & 92.95 & 97.34  & 94.23 & 90.61 & 82.33 & 57.63 \\
ConvBERT+MLM$_{adapt}$  & 87.42 & 94.01 & 93.24 & 97.40 & 95.55 & 91.12 & 82.64 & 58.01 \\
SPACE-2$^{single}$ & 87.83 & 94.68 & 94.05 & 97.56 & 95.58 & 91.20 & 82.66 & 59.08 \\
SPACE-2$^{multi}$ & \textbf{88.10} & \textbf{94.77} & \textbf{94.33} & \textbf{97.80} & \textbf{96.20} & \textbf{91.38} & \textbf{82.74} & \textbf{59.51} \\
\hline 
\ \ \ \ w/o \textit{tree} & 87.57 & 94.19 & 93.49 & 97.44  &96.08 & 91.04 & 82.23 & 58.52 \\
\ \ \ \  w/o \textit{annotation} & 87.33 & 93.93 & 93.23 & 97.33  & 95.99 & 90.93 & 82.02 & 57.87 \\
\ \ \ \ w/o \textit{UnPreDial} & 87.77 & 94.35 & 93.77 & 97.64  & 96.14 & 91.18 & 82.30 & 58.99 \\
\hline
\multicolumn{9}{c}{10-shot data training} \\ \hline
Model & \textit{average}  & \banking & \hwu & \clinic &   \restaurant   & DSTC8  & TOP & \multiWOZ \\ \hline
BERT$^\dagger$ & 66.07 & 79.87 & 81.69 & 89.52 & 87.28 & 45.05 & 74.38 & 4.69 \\
ConvBERT$^\dagger$ & 68.03 &83.63& 83.77& 92.10& 86.90 &49.08 &74.86 &5.90 \\
ConvBERT+MLM$^\dagger$ & 68.22 & 83.99 & 84.52 & 92.75 & 86.17 & 48.40 & 78.84 & 6.87 \\
ConvBERT-DG$^\dagger$ & 73.75 & 84.42 & 85.17 & 92.87 & 87.65 & 41.94 & 75.27 & 48.94 \\
ConvBERT-DG+MLM$^\dagger$ & 73.80 & 85.06 & 85.69 & 93.06 & 87.58 & 44.36 & 72.01 & 48.89 \\
TOD-BERT$_{adapt}$   & 75.70 & 85.99 & 86.74 & 93.07 & 87.62 & 50.19 & 77.77 & 48.54 \\
ConvBERT+MLM$_{adapt}$  & 76.51 & 86.73 & 87.29 & 93.65  & 87.72 & 52.48 & 78.88 & 48.85 \\
SPACE-2$^{single}$ & 77.43 & 88.08 & 88.29 & 94.80 & 88.52 & 53.27 & 79.01 & 50.07 \\
SPACE-2$^{multi}$ & \textbf{77.98} & \textbf{88.31} & \textbf{88.85} & \textbf{95.22} & \textbf{88.85} & \textbf{54.41} & \textbf{79.55} & \textbf{50.70} \\        
\hline 
\ \ \ \ w/o \textit{tree} & 77.31 & 87.41 & 87.55 & 94.76  & 88.30 & 54.26 & 78.95 & 49.97 \\
\ \ \ \  w/o \textit{annotation} & 77.20  & 87.60 & 87.64 & 94.51 & 87.93 & 54.01 & 79.00 & 49.73 \\
\ \ \ \ w/o \textit{UnPreDial} & 77.65 & 88.15 & 88.38 & 94.71  & 88.61 & 54.12 & 79.33 & 50.28 \\
\bottomrule
\end{tabular}
}
\caption{Total results of seven datasets from the DialoGLUE benchmark. $^\dagger$ denotes original results from DialoGLUE.  $_{adapt}$ denotes re-implementation via adaptively pre-training on our corpora. `w/o' denotes ablation study in Sec \ref{sec: ablation}}
\label{tab:dialoglue}
\end{table*}

\subsection{Baselines}
Our compared baselines can be divided into two categories. The first type is the pre-trained model adapted for general dialog understanding tasks in DialoGLUE. We choose BERT \cite{devlin-etal-2019-bert}, ConvBERT \cite{mehri2020dialoglue} and its variants, and TOD-BERT \cite{wu-etal-2020-tod} as our baselines. The second type is the pre-trained model  designed only for specific dialog tasks. In this paper, we choose Example+Observer \cite{mehri-eric-2021-example}, DNNC \cite{zhang-etal-2020-discriminative}, CPFT \cite{zhang2021few}; and ConvFit \cite{vulic2021convfit} for the intent prediction task. We also choose Span-ConveRT \cite{coope-etal-2020-span}, ConVEx \cite{henderson-vulic-2021-convex} and GenSF \cite{mehri-eskenazi-2021-gensf} for the slot-filling task. More details about baselines can be found in Appendix \ref{sec: baselines}. 

\subsection{Settings}
In our experiment, we initialize SPACE-2 with ConvBERT.
The input length is bound to 256 and the batch size is 128. AdamW is used for optimization with an initial learning rate of 1e-5. 
The dropout rate is 0.2.
For semi-supervised pre-training, we combine the labeled and unlabeled data to form a batch in a 1:1 ratio.
For each few-shot experiment, we \textbf{exclude} the  training data in the \textit{AnPreDial} accordingly to avoid unfair data use. For instance, if the target few-shot task is \hwu,  we only use the few-shot training data of \hwu and all other datasets as pre-training resources, so no extra training data (dev\&test) in \hwu is used.
For all downstream tasks, we average 5 seeds for final results.

\section{Experimental Results}
\subsection{Overall Performance on DialoGLUE}
We follow the original settings in \citet{mehri2020dialoglue} and obtain complete results for our model with single-view scoring objectives, denoted as SPACE-2$^{single}$, and with multi-view scoring objectives, denoted as SPACE-2$^{multi}$,  on all seven datasets of DialoGLUE. 
As shown in Table \ref{tab:dialoglue}, our SPACE-2$^{multi}$ model achieves state-of-the-art performance across all tasks, improving the average score of previous best results by \textbf{1.39\%} (86.89\%$\rightarrow$88.10\%) and \textbf{5.66\%} (73.80\%$\rightarrow$77.98\%) on full-data training and 10-shot training settings, respectively. This indicates that our semi-supervised contrastive learning can acquire better pre-trained representations to adapt efficiently to downstream dialog tasks by utilizing both labeled and unlabeled data. 
Although the superior results of SPACE-2$^{single}$ over other baselines, it is worse than SPACE-2$^{multi}$, suggesting that multi-view scoring is crucial to improve the overall performance. 
To eliminate the discrepancy brought by the pre-training dialog corpora, we choose two similar models, TOD-BERT and ConvBERT+MLM, to continually pre-train on  \textit{AnPreDial} and \textit{UnPreDial} for fairness, but they are still not comparable to SPACE-2. 



\begin{figure*}[t]
    \centering
    \includegraphics[width=1.0\textwidth]{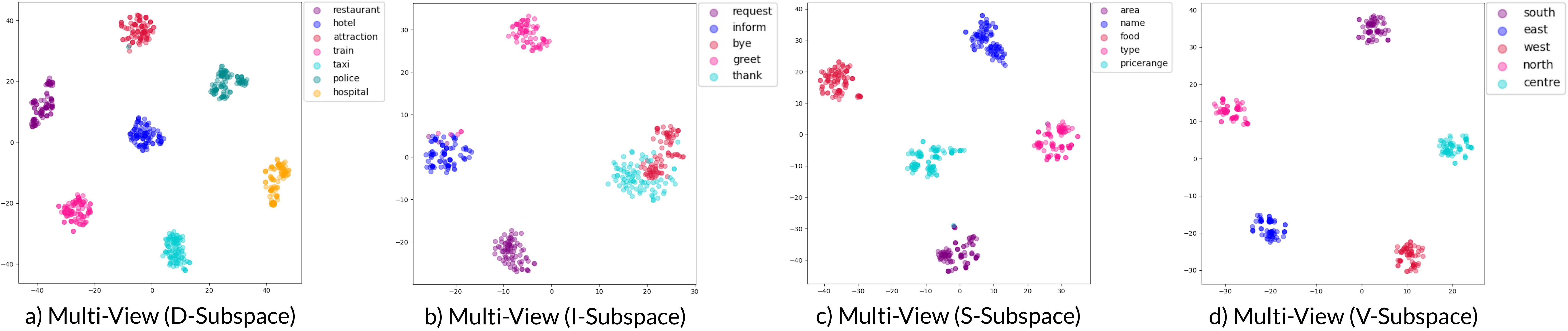}
    \caption{T-SNE visualization of the $\sigma_k(z)$ in multi-view scoring function for \multiWOZ test dialog samples. Four sub-spaces, including Domain-, Intent-, Slot- and Value- Subspace, are shown here.}
    \label{fig:visualize}
\end{figure*}

\subsection{More Comparison on Intent Prediction}

The experimental results of three intent prediction datasets are shown in Table \ref{tab:intent_predict}.  As we can see,  SPACE-2$^{multi}$ achieves SOTA results on all datasets under all settings.  Specifically,
it outperforms the previous best model ConvFiT by 0.93\%, 2.33\% 3.53\% on \banking, \clinic, \hwu under the few-data training setting, respectively. It also improves ConvFiT by 0.61\%, 0.46\%, and 1.91\% on the above three datasets under the full-shot training setting.  The improvements indicate that our method has a better ability to discriminate similar intents via tree-structure enhanced contrastive learning than vanilla pre-training on ConvFiT. 

\subsection{More Comparison on Slot-Filling}
We also conduct slot-filling experiments, particularly in the few-shot setting, to probe whether our model is also suitable for slot extraction. We follow the same data split and settings in GenSF \cite{mehri-eskenazi-2021-gensf} for a fair comparison. As shown in Table \ref{tab:slot_filling}, SPACE-2$^{multi}$ achieves state-of-the-art results across all experimental settings
on DSTC8 and \restaurant.
Note that in \citet{mehri-eskenazi-2021-gensf}, DSTC8 is evaluated in every single domain (Bus, Event, Home, RentalCar) and \restaurant is evaluated with varied proportions (1/8, 1/16, 1/32) of the training set.
It shows that our SPACE-2 model can achieve superior results without any specific designs only for slot-filling tasks.

 \begin{table}[t]
\centering
\resizebox{0.48\textwidth}{!}{
\begin{tabular}{lcccccc}
\toprule
\multirow{2}{*}{Model} & \multicolumn{2}{c}{\banking} & \multicolumn{2}{c}{\clinic} & \multicolumn{2}{c}{\hwu} \\
 & Few & Full & Few & Full & Few & Full \\
 \midrule
Example+Observer  & 85.95 & 93.83 & 93.97 & 97.31 & 86.28 & 93.03 \\
DNNC  & 86.71 & - & 93.76 & - & 84.72 & - \\
CPFT  & 87.20 & - & 94.18 & - & 87.13 & - \\
ConvFiT & 87.38 & 94.16 & 92.89 & 97.34 & 85.32 & 92.42 \\
SPACE-2$^{multi}$ & \textbf{88.31} & \textbf{94.77} & \textbf{95.22} & \textbf{97.80} & \textbf{88.85} & \textbf{94.33}\\
\bottomrule
\end{tabular}}
\caption{Results on intent prediction  tasks, including 10-shot training (Few) and full-set training (Full) settings.}
\label{tab:intent_predict}
\end{table}



\subsection{Ablation Study}
\label{sec: ablation}
To figure out which factor contributes mainly to our semi-supervised pre-training paradigm, we conduct ablation studies on DialoGLUE as shown in the bottom part of each setting in Table \ref{tab:dialoglue}.  
We first investigate the effect of the tree-structured supervised contrastive learning by replacing it with the traditional supervised contrastive learning (denote as `w/o \textit{tree}'). As we can see, vanilla supervised contrastive learning only considers self-augmented data or data with the exact same labels as positive samples, so it can not capture a more detailed semantic structure than our tree-structured method and degrade from 88.10\% to 87.57\% in an average score.
We also try to remove the label information from \textit{AnPreDial} and treat all pre-training data as unlabeled data  (denote as `w/o \textit{annotation}'), or remove the unlabeled data  (denote as `w/o \textit{UnPreDial}') to see the impact.
It shows that both the labeled data and unlabeled data are crucial in our semi-supervised pre-training scheme but labeled data contributes more to the overall performance.

\subsection{Visualization Analysis}
\label{sec: analysis}
In this part, we investigate why explicitly decomposing the semantic structure into sub-structures for different views can benefit dialog understanding. 
Figure \ref{fig:visualize} illustrates the 2D t-SNE visualization of the output unit vectors $\sigma_k(z)$ for test dialog samples from the  \multiWOZ dataset. Due to the limited space, we only show the sub-spaces of  \texttt{D,I,S,V} here. As we can see, 
The hidden representations of SPACE-2$^{multi}$ are able to differentiate the similar and dissimilar parts in different semantic sub-spaces.
The learned latent sub-space is highly correlated with the dialog annotations of \textit{domain, intent, slot} or \textit{value}, which confirms our assumption.




\begin{table}[t]
\centering
\resizebox{0.48\textwidth}{!}{
\begin{tabular}{lcccc}
\toprule
Dataset     & Span-ConveRT & ConVEx & GenSF & SPACE-2$^{multi}$ \\ \hline
DSTC8-Bus      & 84.0         & 86.7   & 90.5  &  \textbf{91.6}     \\
DSTC8-Event     & 82.2         & 87.2   & 91.2  &  \textbf{92.4}     \\
DSTC8-Home      & 95.4         & 94.5   & 93.7  &   \textbf{96.2}    \\
DSTC8-RentalCar & 83.0         & 87.4   & 86.7  &   \textbf{88.3}    \\ 
\midrule
\restaurant-1/8  & 88.5         & 90.6   & 91.8  &  \textbf{93.9}    \\
\restaurant-1/16 & 81.1         & 86.4   & 89.7  &  \textbf{92.6}    \\ 
\restaurant-1/32 & 63.8         & 81.8   & 82.1  &  \textbf{84.7}     \\
\bottomrule
\end{tabular}
}
\caption{F1 scores across all slots for evaluation on the DSTC8 single-domain datasets in the few-shot setting.}
\label{tab:slot_filling}
\end{table}

\section{Conclusion}
\label{sec:conclu}
In this work, we propose a new pre-trained conversation model named SPACE-2,  which learns dialog representations from both labeled and unlabeled corpora via tree-structured semi-supervised contrastive learning (CL). 
To be specific, we utilize a semantic tree structure (STS) to unify the inconsistent annotation schema for all datasets. Then, we propose a novel multi-view scoring function to compute the semantic similarity based on STS's sub-structures for labeled dialogs. Different from  traditional CL, our method increases the relevance of representations for all similarly labeled dialogs and only pushes away completely different dialogs. 
A self-supervised CL loss is also integrated to exploit unlabeled dialogs.
Extensive experiments on the DialoGLUE benchmark with seven datasets demonstrate that our model achieves the best results on all dialog understanding tasks, under the full-data and the few-shot settings.

\section*{Acknowledgement}
This work was partially supported by National Natural Science Foundation of China (No. 61906185), Youth Innovation Promotion Association of CAS China (No. 2020357),
Shenzhen Science and Technology Innovation Program (No. KQTD20190929172835662), Shenzhen Basic Research Foundation (No. JCYJ20210324115614039 and No. JCYJ20200109113441941).
This work was supported by Alibaba Group through Alibaba Research Intern Program.

\bibliographystyle{acl_natbib}
\bibliography{anthology,acl2021}

\clearpage
\appendix
\section{Details for Pre-training Dialog Corpora}
\label{sec: appen}

All dialog corpora we used in our experiments are given in Table \ref{tab:LabelPreDialstatictis} and Table \ref{tab:UnlabelPreDialstatictis}. There are totally 32 labeled datasets and 21 unlabeled datasets. 
For the labeled data, each dialog sample is unified into the same data format with tree-structured labels.
For the unlabeled data, each sample is performed careful cleaning due to the high noise. The processing methods include: 
(1) removing the instances where there is a URL in utterances. 
(2) removing the instances containing word repetitions of at least three words.
(3) removing non-English sentences. 
(4) removing sentences containing special markers such as “[” or “]”, as this could be markup. 
(5) removing offensive language.
(6) replacing the non-unicode characters like emojis.

\section{Input Representations}
\label{sec: input_representations}

The input embeddings consist of four elements: tokens, roles, turns, and positions. Role embeddings, $E_u, E_s$, are like segmentation embeddings in BERT and are used to differentiate which role the current token belongs to either user or system. Turn embeddings are assigned to each token according to its adverse turn number in the input dialog data. Position embeddings are assigned to each token according
to its relative position within its belonging sentence. For the tokens, we use special tokens \texttt{[BOU]}, \texttt{[EOU]} to bound the user utterance, and \texttt{[BOS]}, \texttt{[EOS]} to bound the system response. 

\section{Details for Compared Baselines}
\label{sec: baselines}
\subsection{Baselines for Dialog Understanding}
We choose the following baselines: 1) \textbf{BERT}, a BERT$_{\text{base}}$ model adapted to all downstream tasks; 2) \textbf{ConvBERT}, a BERT continually pre-trained on a large open-domain dialogue corpus with MLM  \cite{mehri2020dialoglue}; 3) \textbf{ConvBERT-DG}, a ConvBERT  continually pre-trained  on all DialoGLUE data before conducting specific downstream tasks with MLM. 
4) \textbf{ConvBERT+MLM}, a ConvBERT  continually trained on training data of specific downstream tasks; 5) \textbf{ConvBERT-DG+MLM}, a ConvBERT+MLM pre-trained on all DialoGLUE data before conducting downstream tasks; and 6) \textbf{TOD-BERT}, a BERT pre-trained on nine existing task-oriented dialogue corpora using MLM and response selection loss \cite{wu-etal-2020-tod}.







\subsection{Baselines for Specific Tasks}
In this paper, we focus on intent prediction and slot-filling in particular since there are adequate pre-trained baselines specifically.
For the intent prediction task, we choose: 1) \textbf{Example+Observer}, a ConvBERT+MLM that uses example-driven training based on similarity matching and observers for transformer attentions \cite{mehri-eric-2021-example}; 2) \textbf{USE+ConveRT}, a dual sentence encoder pre-trained on 654M Reddit dialogs \cite{casanueva-etal-2020-efficient}; 3) \textbf{DNNC}, a discriminative nearest-neighbor model which finds the best-matched sample via similarity matching \cite{zhang-etal-2020-discriminative}; 4) \textbf{CPFT}, which uses self-supervised contrastive learning to perform  pre-training and fine-tuning  \cite{zhang2021few}; and 5) \textbf{ConvFit}, which uses regular supervised contrastive learning on labeled data after pre-training on chatting data \cite{vulic2021convfit}. 
For the slot-filling task, we choose: 1)  \textbf{Span-ConveRT}, a slot-filling model pre-trained on collected data 
\cite{coope-etal-2020-span}; 2) \textbf{ConVEx}, a span-ConveRT pre-trained with the pairwise cloze task on Reddit for few-shot slot labeling \cite{henderson-vulic-2021-convex}; and 3) \textbf{GenSF}, a generative pre-trained slot-filling model \cite{mehri-eskenazi-2021-gensf}.

\begin{table*}[hbp]
\centering
\scalebox{1.0}{
\begin{tabular}{lrr}
\hline
Dataset                   & \# Dialog & \# Utterance \\ \hline
\banking\cite{casanueva-etal-2020-efficient}              & 13,083    & 13,083       \\
\clinic\cite{larson-etal-2019-evaluation}               & 23,700    & 23,700       \\
\hwu\cite{liu2021benchmarking}                  & 25,716    & 25,716       \\
\restaurant\cite{coope-etal-2020-span}                 & 11,115    & 11,975       \\
TOP\cite{gupta-etal-2018-semantic-parsing}                    & 44,783    & 44,783       \\
ATIS\cite{hemphill-etal-1990-atis}                   & 5,817     & 5,817        \\
SNIPS\cite{coucke2018snips}                  & 14,484    & 14,484       \\
CrossNER\cite{liu2020crossner}               & 27,458    & 27,458       \\
FB\_TOD\_SF\cite{schuster-etal-2019-cross-lingual}         & 43,323    & 43,323       \\
MIT-restaurant\cite{liu2013asgard}         & 9,181     & 9,181        \\
MIT-movies-eng\cite{liu2013asgard}         & 12,218    & 12,218       \\
MIT-movies-trival10k13\cite{liu2013asgard} & 9,769     & 9,769        \\
\multiWOZ\_coco\cite{li2020coco}         & 32,062    & 64,124  \\
\multiWOZ\cite{eric-etal-2020-multiwoz}               & 10,433    & 142,968      \\
STAR\cite{mosig2020star}                   & 5,820     & 98,962       \\
DailyDialog\cite{li2017dailydialog}            & 13,118    & 102,979      \\
SGD\cite{rastogi2020towards}                    & 22,825    & 463,284      \\
Frames\cite{el-asri-etal-2017-frames}                 & 1,369     & 19,986       \\
MSRe2e\cite{li2018microsoft}                 & 10,087    & 74,686       \\
DSTC2\cite{williams2016dialog}                  & 4,953     & 73,228       \\
DSTC3\cite{williams2016dialog}                  & 1,980     & 30,456       \\
SimJoint\cite{shah-etal-2018-bootstrapping}               & 3,008     & 27,120       \\
MulDoGo\cite{peskov-etal-2019-multi}                & 11,668    & 194,097      \\
WOZ\cite{mrksic-etal-2017-neural}                    & 1,200     & 8,824        \\
TaskMaster1\cite{byrne-etal-2019-taskmaster}            & 13,210    & 279,287      \\
TaskMaster2\cite{byrne-etal-2019-taskmaster}            & 17,289    & 292,830      \\
TaskMaster3\cite{byrne-etal-2019-taskmaster}            & 23,757    & 477,026      \\
InCar\cite{eric-etal-2017-key}                  & 3,031     & 15,928       \\
\multiWOZ\_synthesis\cite{campagna-etal-2020-zero}    & 37,605    & 401,075      \\
SwDA\cite{stolcke-etal-2000-dialogue}                   & 1,434     & 274,786      \\
BiTOD\cite{lin2021bitod}                  & 3,689     & 72,462       \\
PersuaGOOD\cite{wang-etal-2019-persuasion}             & 300       & 10,864       \\ \hline
Total                  & 459,465   & 3,366,479    \\ \hline
\end{tabular}
}
\caption{Statistics for each labeled dataset in \textit{AnPreDial}.}
\label{tab:LabelPreDialstatictis}
\end{table*}

\begin{table*}[htp]
\centering
\resizebox{0.7\textwidth}{!}{
\begin{tabular}{lrr}
\hline
Name          & \# Dialog  & \# Utterance \\ \hline
MulDoGo\_un\cite{peskov-etal-2019-multi}   & 63,404     & 1,013,985      \\
ABCD\cite{chen2021action}          & 10,042     & 143,855       \\
AirDialog\cite{wei-etal-2018-airdialogue}     & 361,822    & 4,645,475      \\
CCPE\cite{radlinski-etal-2019-coached}          & 502        & 11,971        \\
MetalWOZ\cite{shalyminov2020fast}      & 40,203     & 458,237       \\
CMU\_DoG\cite{zhou-etal-2018-dataset}      & 4,221      & 134,197       \\
ConvQuestions\cite{kacupaj-etal-2021-conversational} & 11,200     & 112,000       \\
CoQA\cite{reddy-etal-2019-coqa}          & 7,699      & 233,260       \\
CoSQL\cite{yu-etal-2019-cosql}         & 2,458      & 25,433        \\
doc2dial\cite{feng-etal-2020-doc2dial}      & 688        & 52,688        \\
DSTC10-track2\cite{kim2021how} & 2,919      & 55,377        \\
DSTC10-track3\cite{kottur2021simmc} & 11,244     & 110,767       \\
MedicalDialog\cite{zeng-etal-2020-meddialog} & 482        & 981          \\
Self-Dialog\cite{fainberg2018talking}   & 24,165     & 372,720       \\
WOW\cite{dinan2018wizard}           & 22,311     & 201,999       \\
TopicChat\cite{gopalakrishnan2019topical}     & 8,628      & 188,378       \\
Persona-Chat\cite{zhang-etal-2018-personalizing}  & 11,087     & 81,032     \\ 
MMD\cite{1704.00200}           & 1,506,129  & 5,477,066      \\
CSQA\cite{1801.10314}          & 197,001    & 4,153,092      \\
AmazonQA\cite{gupta2019amazonqa}      & 923,685    & 1,847,370      \\
ChitChat\cite{will2020conversational}      & 7,168      & 258,145       \\ \hline
Total         & 3,217,058  & 19,578,028   \\ \hline
\end{tabular}
}
\caption{Statistics for each unlabeled dataset in \textit{UnPreDial}.}
\label{tab:UnlabelPreDialstatictis}
\end{table*}


\end{document}